\documentclass[3p,times,twocolumn,longtitle,sort&compress, authoryear]{elsarticle}
\usepackage{ecrc}
\usepackage{natbib}
\usepackage{cite}
\usepackage{graphicx}
\usepackage{float}
\usepackage{subfig}
\usepackage{booktabs}
\usepackage{multirow}
\usepackage{amsmath,amsfonts}
\usepackage{color}

\volume{00}
\firstpage{1}
\journalname{Expert Systems With Application}
\runauth{J. Qian et al.}
\jid{procs}
\jnltitlelogo{}
\CopyrightLine{2021}{Published by Elsevier Ltd.}

\usepackage{amssymb}
\usepackage[figuresright]{rotating}

\begin{document}

\begin{frontmatter}

\dochead{}

\title{HASA: Hybrid Architecture Search with Aggregation Strategy for Echinococcosis Classification and Ovary Segmentation in Ultrasound Images}

\author[a1,a2,a3]{Jikuan Qian\fnref{fn1}}
\ead{beiahi.qian@gmail.com}

\author[a1,a2,a3]{Rui Li\fnref{fn1}}
\ead{945193029@qq.com}

\author[a1,a2,a3]{Xin Yang}
\ead{yangxinknow@gmail.com}

\author[a1,a2,a3]{Yuhao Huang}
\ead{huangyuhao2019@email.szu.edu.cn}

\author[a1,a2,a3]{Mingyuan Luo}
\ead{luomingyuan2020@email.szu.edu.cn}

\author[a1,a2,a3]{Zehui Lin}
\ead{linzehui2020@email.szu.edu.cn}

\author[a1,a2,a3]{Wenhui Hong}
\ead{hwh51588@qq.com}

\author[a1,a2,a3]{Ruobing Huang}
\ead{ruobing.huang@szu.edu.cn}

\author[a4]{Haining Fan}
\ead{1486713174@qq.com}

\author[a1,a2,a3]{Dong Ni}
\ead{nidong@szu.edu.cn}

\author[a1,a2,a3]{Jun Cheng\corref{cor1}}
\ead{juncheng@szu.edu.cn}

\address[a1]{National-Regional Key Technology Engineering Laboratory for Medical Ultrasound, School of Biomedical Engineering, Health Science Center, Shenzhen University, Shenzhen, China}
\address[a2]{Medical Ultrasound Image Computing (MUSIC) Laboratory, Shenzhen University, Shenzhen, China}
\address[a3]{Marshall Laboratory of Biomedical Engineering, Shenzhen University, Shenzhen, China}
\address[a4]{Qinghai University Affiliated Hospital, Xining, Qinghai, China}

\cortext[cor1]{Corresponding author}
\fntext[fn1]{Jikuan Qian and Rui Li contributed equally to this work.}

\begin{abstract}
Different from handcrafted features, deep neural networks can automatically learn task-specific features from data. Due to this data-driven nature, they have achieved remarkable success in various areas. However, manual design and selection of suitable network architectures are time-consuming and require substantial effort of human experts. To address this problem, researchers have proposed neural architecture search (NAS) algorithms which can automatically generate network architectures but suffer from heavy computational cost and instability if searching from scratch. In this paper, we propose a hybrid NAS framework for ultrasound (US) image classification and segmentation. The hybrid framework consists of a pre-trained backbone and several searched cells (i.e., network building blocks), which takes advantage of the strengths of both NAS and the expert knowledge from existing convolutional neural networks. Specifically, two effective and lightweight operations, a mixed depth-wise convolution operator and a squeeze-and-excitation block, are introduced into the candidate operations to enhance the variety and capacity of the searched cells. These two operations not only decrease model parameters but also boost network performance. Moreover, we propose a re-aggregation strategy for the searched cells, aiming to further improve the performance for different vision tasks. We tested our method on two large US image datasets, including a  9-class echinococcosis dataset containing 9566 images for classification  and an ovary dataset containing 3204 images for segmentation. Ablation experiments and comparison with other handcrafted or automatically searched architectures demonstrate that our method can generate more powerful and lightweight models for the above US image classification and segmentation tasks.
\end{abstract}


\end{frontmatter}

%% main text
\section{Introduction}
\label{sec:introduction}
Ultrasound (US) imaging is preferred for many clinical examinations for its advantages of being real-time, non-invasive and radiation-free. It has become one of the most commonly performed imaging modalities in clinical practice. Periodical and comprehensive ultrasound screenings are usually performed to monitor the status of lesion area.  However, due to the poor image quality and the presence of speckles and acoustic shadows, interpreting  US images is a highly skilled task and suffers from large inter- and intra-observer variations. From the perspective of clinical practices, it is highly demanded to develop advanced automatic US image analysis methods for objective clinical diagnosis and assessment \citep{liu2019deep}. \par

Compared to the traditional feature engineering methods, deep neural networks (DNNs) can directly learn the discriminative features from data in an end-to-end way \citep{skinClass,miahi2021genetic}. Such methods have proved to be an efficient and powerful tool for many automatic US image analysis tasks, such as object detection \citep{detection1,detection2,zhou20213d} , target segmentation \citep{seg1,seg2,singh2020breast}, and image classification \citep{class1,class2}. In order to obtain good performance, various DNN architectures tailored for specific tasks are devised such as the ResNet \citep{he2016deep}, DenseNet \citep{dense2017}, and EfficientNet \citep{efficient2019}, and a large dataset needs to be curated to avoid model overfitting \citep{skinClass}.  \par

However, due to the inherent artifacts of US images and the challenges of tasks, it is tricky to manually design effective DNN architectures for automatic analysis. For example, the performance of US image segmentation is often influenced by the characteristic artifacts such as attenuation, speckles, shadows, and multi-scale variations of objects. Although the multi-scale fusion design, such as U-Net\citep{ronneberger2015u}, improves the situation by utilizing both high-level semantic features and low-level features, an important factor that still limits the performance may be the handcrafted architecture itself.  For classification tasks, the power of learning discriminative high-level features is critical for achieving good performance. Due to the high inter-class similarity between disease subtypes in US images, a well-designed network architecture is required to fully distill the class-specific representations. However, in practice, designing a network architecture with excellent performance not only requires a profound knowledge of the dataset and the mechanism of the learning process, but also costs a lot of efforts to perform extensive experiments for the hyper-parameter selection. \par

To address the  aforementioned problems, researchers have proposed neural architecture search (NAS) algorithms that can automatically generate effective network architectures. NAS first obtains a search space where all possible network architectures or network components are defined. Next, it carries on a search strategy to propose an architecture candidate from the search space. NAS evaluates the candidate based on an estimation strategy. The feedback is then used to update the architecture and model parameters. The process is repeated iteratively until convergence. Such a strategy can generate network architectures automatically.  Although early works on NAS achieved great performance, the computational cost was unacceptably high due to the huge search space and large memory cost \citep{zoph2016neural,real2017large}. Moreover, to find the optimal network in a large search space, most NAS methods require a large amount of training data to guide the search. When only a small dataset is available, the search performance might be degraded.  \par

In this work, to address the above issues, we propose a new hybrid architecture search (HAS) framework which incorporates expert knowledge from existing neural networks into NAS to improve the model search performance. The HAS framework is fast in searching (about 10 GPU hours), memory-efficient, and well-adapted to small datasets. We further equip the HAS with a cell re-aggregation strategy to get an enhanced version, HASA. A cell is a computing structure represented by a directed acyclic graph (see Section 3.3). The cell re-aggregation strategy utilizes expert knowledge to transform the searched cells into task-specific hierarchies to get better performance. \par

\begin{figure}[t]
	\centerline{\includegraphics[width=\columnwidth]{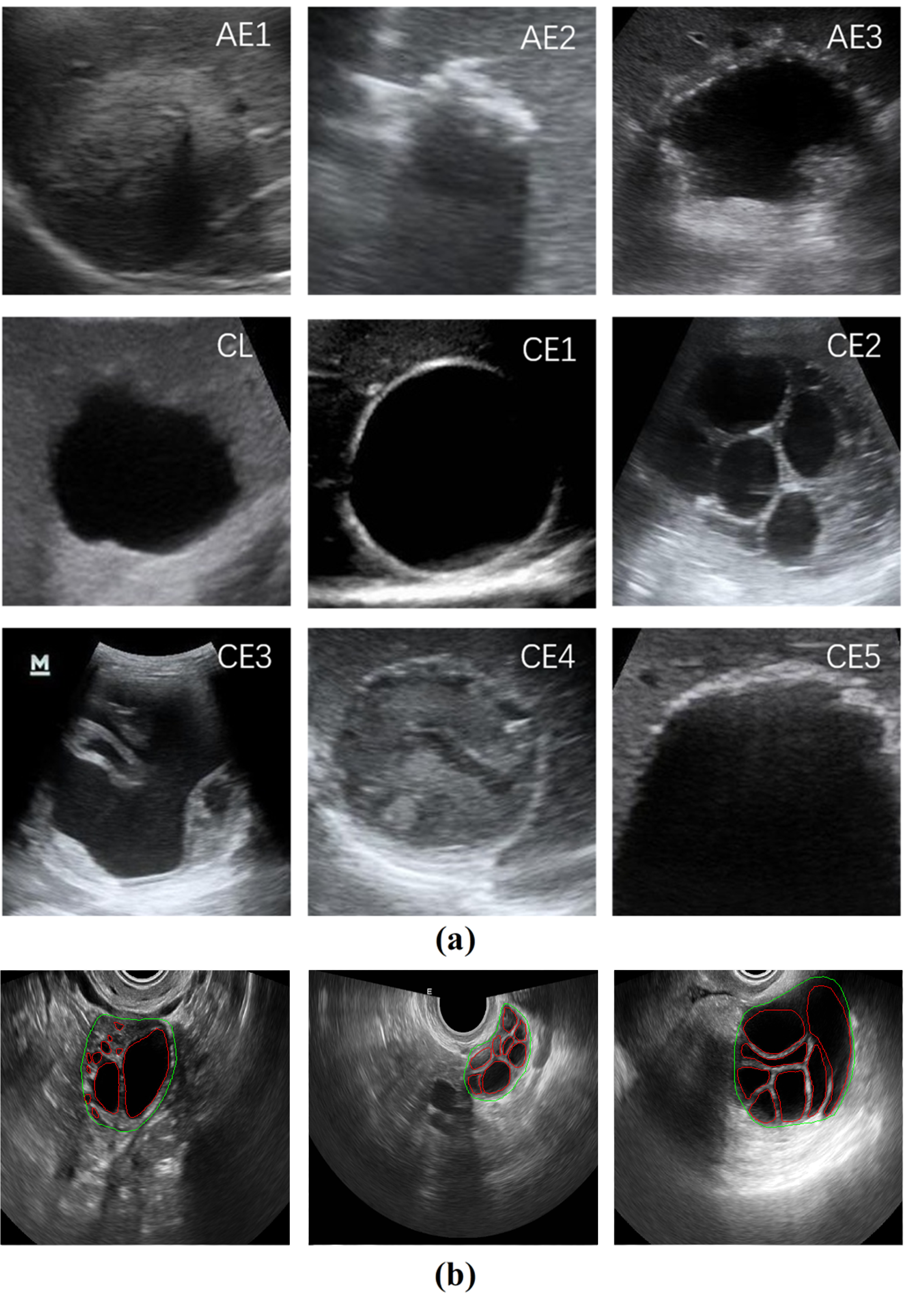}}
	\caption{The typical ultrasound images of echinococcosis with 9 classes (a) and ovary and follicles (b). The green and red contours represent the ovary and follicles, respectively.}
	\label{example_pic}
\end{figure}

To summarize, our main contributions are highlighted as follows:
\begin{enumerate}[1)]
    \item 
    Instead of starting from scratch to search for the whole network, our proposed hybrid framework combines a pre-trained backbone and a series of NAS cells. In this way, the search space can be large narrowed. As in transfer learning, the use of a pre-trained backbone can effectively improve model performance on a small dataset and accelerate model convergence during training.
    \item
    We adopt a progressive growing strategy that breaks the entire search space into smaller spaces, starting with a simple model and growing to more complex ones. This strategy can stabilize the search process and efficiently identify the most promising model.
    \item
    We introduce a mixed depth-wise convolution operator and a squeeze-and-excitation block into the candidate operations, which enriches the variety of cell structure and thus enables discovering more powerful cells. After the cell search process, we re-aggregate the sequentially stacked cells into the expert-designed structures to further enhance the model performance for different tasks.
    \item
    We conduct experiments on two large US image datasets: a 9-class echinococcosis dataset with 9566 images for the classification task and an ovary dataset with 3204 images for the segmentation task. Example images of the two datasets are shown in Fig.~\ref{example_pic}. The superior performance over other handcrafted and automatically searched models and ablation studies demonstrate the efficacy of our method.
\end{enumerate}

The remainder of this paper is organized as follows. Section 2 summarizes related research on NAS as well as manually designed models for image classification and segmentation. Section 3 describes in detail the proposed method. In Section 4, the implementation and test results for image classification and segmentation are reported. Finally, discussion and conclusions are presented in Section 5.

\section{Related work}
The related researches of NAS can be traced back to a dozen years ago, even before the emergence of interest in deep learning. Early search approaches mainly involved random search, grid search \citep{randomSearch}, and simple evolutionary algorithms \citep{evolution1,evolution2}. They are applied to find a proper setting in a small search space (e.g. a fully connected neural network). With the development of deep learning and computing power, the size of neural networks is increased to address more difficult tasks, which also brings huge challenges for the NAS research, such as longer training time and larger search space. To overcome these issues, many research groups have been devoted to more efficient NAS algorithms, which can be classified into three categories: reinforcement learning (RL) based methods, evolution learning (EL) based methods, and continuous differentiable methods. NAS methods have been applied to  a wide range of tasks, such as image classification, segmentation, standard plane localization in ultrasound, and object detection. Here, we provide a brief review of the three categories of NAS algorithms as well as manually designed network architectures for image classification and segmentation. \par

\subsection{Reinforcement learning based search}
Several studies have attempted to use RL based methods to determine neural architectures sequentially. \citet{zoph2016neural} first proposed a RL based NAS method that can automatically find the network architecture for image classification and natural language understanding. Following this scheme, they introduced the cell-wise search and achieved state-of-the-art accuracy on CIFAR-10 and ImageNet \citep{zoph2018}. \citet{baker2016RL} also adopted a RL based strategy to automatically generate convolutional neural network architectures and achieved competitive results on several image classification benchmarks. \citet{zhong2018RL} used a block-wise NAS with a Q-learning strategy and achieved state-of-the-art image classification performance on CIFAR-10 and CIFAR-100. \citet{cai2018proxylessnas} proposed ProxylessNAS to directly learn the architectures for large-scale image classification tasks. However, most RL based approaches tend to require massive computation during the search. For example, the previous work \citep{zoph2016neural} took 2000 GPU-days to complete the search process.

\subsection{Evolution learning based search}
An alternative to RL based search is the neuro-evolutionary method. The first such approach was proposed by \citet{firstNAS} who used genetic algorithms to search architectures and used backpropagation to optimize their weights. Since then, many EL based approaches have tended to use genetic algorithms to search for network architectures automatically. \citet{angeline1994EL} used an EL algorithm to construct a recurrent neural network and applied it to several problems from language induction to search and collection. Then, \citet{stanley2002EL} proposed to evolve neural networks based on augmenting topologies. \citet{stanley2009EL} proposed a hypercube-based EL method for evolving large-scale networks and tested this method on two tasks: visual discrimination and robot food gathering. With the development of deep learning, some researchers also tried to use an EL based approach for different image processing tasks. \citet{liu2019auto} designed a hierarchical NAS strategy for semantic image segmentation. \citet{real2019} proposed regularized evolution to build an image classifier. \citet{xie2017genetic} limited the number of nodes in a cell to further reduce the size of the entire search space and demonstrated the effectiveness on several image classification benchmarks. \citet{guo2019single} proposed a Single Path One-Shot (SPOS) model and achieved state-of-the-art performance on the ImageNet dataset. However, such methods still require considerable training time and computing resources.

\subsection{Continuous differentiable search}
Gradient-based methods have also been proposed in recent years. It significantly reduces the training time of the search process. For example, \citet{liu2018darts} proposed the differentiable architecture search (DARTS), which constructed networks by stacking basic network cells. DARTS reduced computation cost to the same order of magnitude as training a single neural network and ourperformed other methods in image classification and language modeling tasks. Based on DARTS, \citet{chen2019progressive} further proposed the progressive differentiable architecture search (P-DARTS). P-DARTS allowed the depth of searched architectures to grow gradually during the training procedure and achieved superior performance on the CIFAR-100 and ImageNet datasets. \citet{kim2019scalable} pre-defined a U-shape network and filled it with basic cells to search for an optimized architecture for 3D medical image segmentation. \citet{bae2019resource} revisited the classic U-Net and proposed a resource-optimized NAS method for 3D medical image segmentation. Combining DARTS and U-Net, \citet{zhu2019v} proposed to search for architectures for CT image segmentation. Nonetheless, their search space was mainly limited to several 2D and 3D vanilla convolutional operators. 

\subsection{Handcrafted architectures for image classification and segmentation}
Manually designed neural architectures without NAS also developed rapidly in recent years. For the classification tasks, since 2015, ResNet~\citep{he2016deep} has achieved great performance due to the design of residual block. Based on the ResNet, DarkNet53 was proposed as the backbone in YOLOv3~\citep{redmon2016yolo}. Squeeze-and-excitation (SE) networks~\citep{hu2018squeeze} were then proposed to achieve channel-level attention of features. It has been validated that the SE-block is a general module to improve the classification frameworks (e.g., ResNetXt~\citep{xie2017aggregated}, one of the variants of ResNet). There are also some other variants of the ResNet, including Res2net~\citep{gao2019res2net}, ResNetSt~\citep{zhang2020resnest}, and NF-ResNet~\citep{brock2021high}. For the segmentation tasks, U-Net~\citep{ronneberger2015u} and DeepLabv3+~\citep{chen2018encoder} are the most commonly used solutions. Due to its powerful performance, U-Net has been recognized as the baseline in many segmentation tasks. The U-Net family, including UNet++~\citep{zhou2019unet++}, nn-Unet~\citep{isensee2019automated}, and UNet3+~\citep{huang2020unet}, were also proposed. \par

Although the above methods show state-of-the-art performance in medical image classification and segmentation tasks, the design of their architecture requires plenty of time and domain knowledge. Besides, the manually designed architecture itself may limit its performance. Thus, NAS methods are highly desirable to obtain models with better performance, less human involvement, and less computational costs. \par

\begin{figure*}[t]
	\centering
	\includegraphics[width=\linewidth]{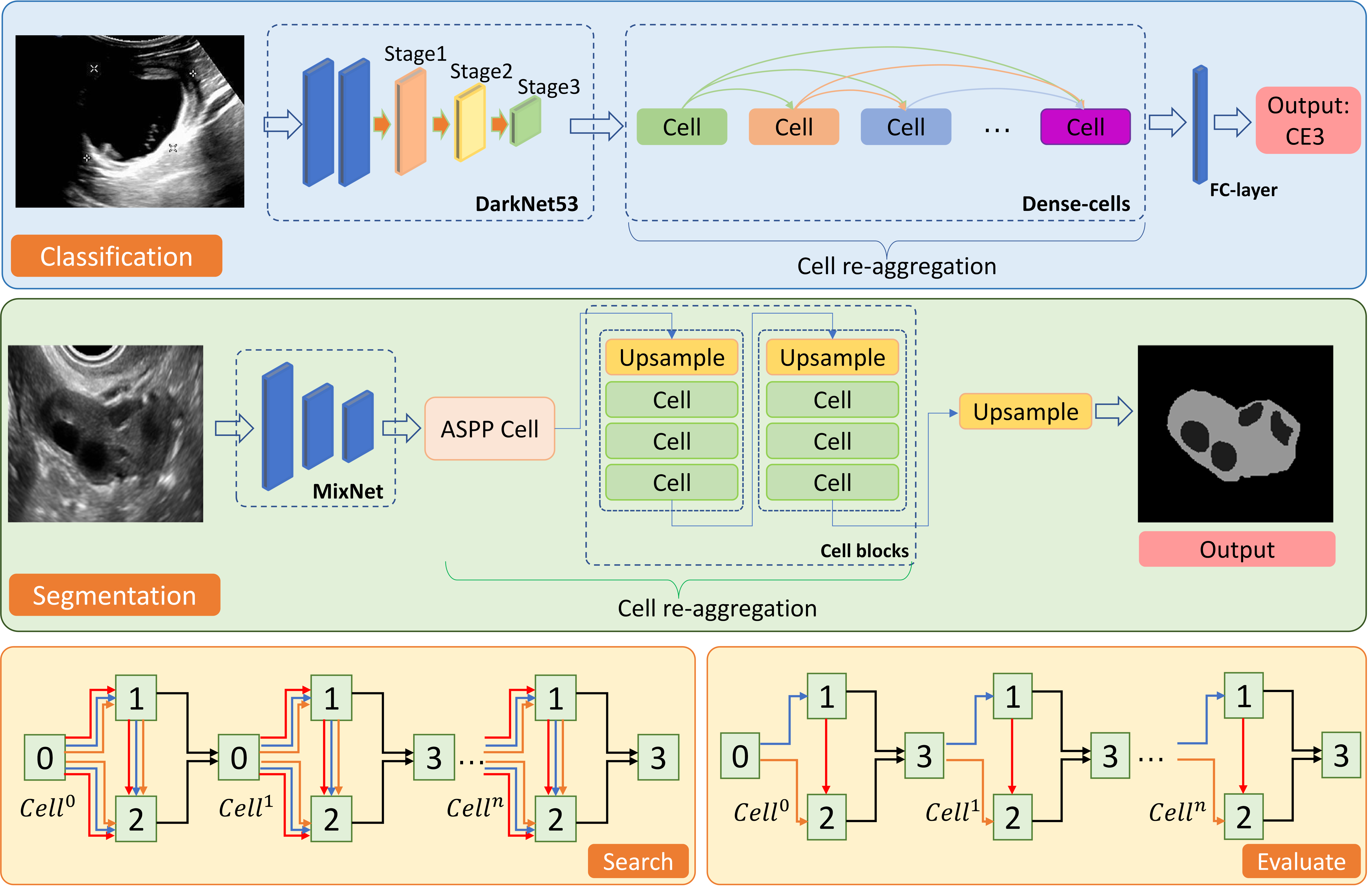}
	% \vspace{-3mm}
	\caption{The schematic diagram of our proposed framework for echinococcosis classification and ovary segmentation. The hybrid framework combines a pre-trained backbone and NAS cells. After the search completes, the sequentially stacked cells are re-aggregated into the Dense-cell structure for the classification task and into the ASPP-cell structure for the segmentation task. Cells in the searching stage with full edges and that in the evaluation stage with selected edges are displayed separately at the bottom.}
	\label{fig:framework}
	% \vspace{-5mm}
\end{figure*}

\section{HASA framework}
We start with an overview of the proposed HASA framework. Next, we describe the backbones used in our framework. We then detail the search process of the stacked cell architecture, including the cell structure design and the progressive growing strategy. Finally, we present how the searched cells are re-aggregated into the expert-designed structures to enhance the model performance for different tasks.

\subsection{Overview of the hybrid framework}
Fig.~\ref{fig:framework} is the overview of our proposed framework. The proposed hybrid network consists of a pre-trained backbone and several NAS cells. The training process has two stages: the cell searching stage and the evaluation stage. We first search the space of cell structures in a progressive way, by learning the operations on its edge through a bi-level optimization. At the evaluation stage, those stacked cells are re-aggregated to form the task-specific hierarchy. The parameters of the model are then fine-tuned to adapt to the new aggregation structures. Finally, we evaluate the model on the test dataset. \par

The hybrid framework involves the idea of transferring learning. We freeze the parameters of shallow layers and  fine tune the parameters of deeper layers to solve task-specific problems. However, as opposed to vanilla transfer learning, we introduce the NAS to the fine-tuning process which results in a hybrid model with a pre-trained backbone and the searched cells. With different types of backbones, we can generalize our method to different tasks (such as classification and segmentation). Featured with the pre-trained backbone, the searching space of NAS is largely reduced, which alleviates heavy computational burden. In addition, transfer learning allows us to leverage existing knowledge and avoid overfitting.

\subsection{Backbones}
For the echinococcosis classification, we use DarkNet53~\citep{redmon2018yolov3} as our backbone. DarkNet53 was proposed as the backbone in YOLOv3. It mainly consists of convolutional layers and residual structures, which have powerful feature learning capability. DarkNet53 replaces the pooling layer with a step-size varying convolutional layer to reduce the information loss in the feature map.

\begin{figure}[htbp]
	\centering
	\includegraphics[width=1.0\linewidth]{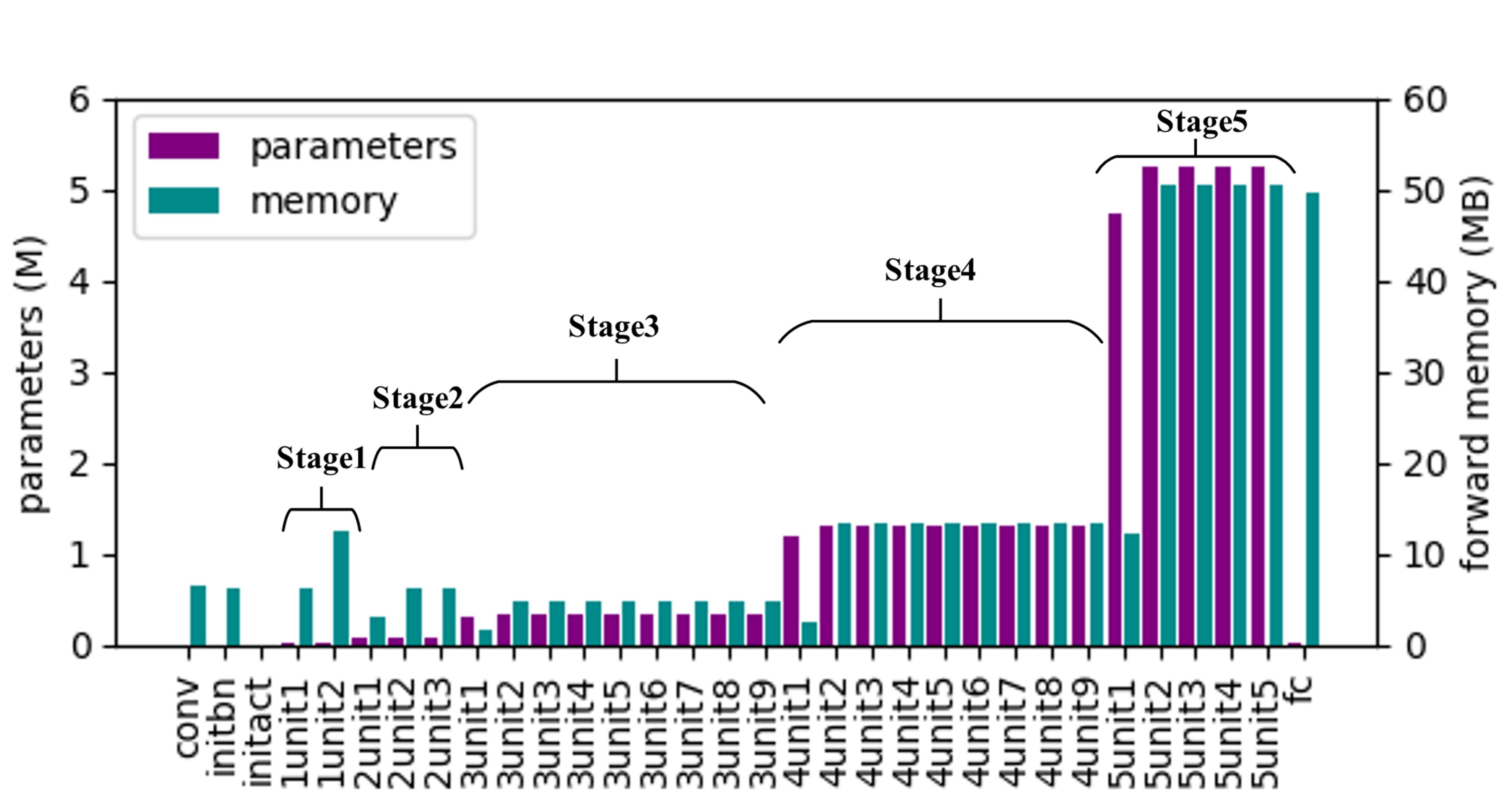}
	\caption{The parameters and memory cost of DarkNet53.}
	\label{memory}
\end{figure}

DarkNet53 contains 5 stages with several residual blocks. Similar to ResNet, the shallow stages are expected to extract texture information while the deep stages are expected to extract semantic information. We assume that deep stages in the DarkNet53 involve more dataset-specific features which are more
desired to customization than the shallow stages. Besides, as Fig.~\ref{memory} shows, the 4$^{th}$ and 5$^{th}$  stages take up major parameters and memory consumption in all stages. Therefore, we replace the 4$^{th}$ and 5$^{th}$ stages with cells and searched these cells for higher efficiency and accuracy. 

For the ovary segmentation, we use MixNet~\citep{tan2019mixconv} as our backbone. MixNet is a lightweight network and features a mixed depthwise convolution. It mixes convolution kernels of different sizes into the same convolution unit. By capturing feature patterns with different resolutions, mixed depthwise convolution can lead to better accuracy and efficiency, compared with using vanilla depthwise convolution.

\subsection{Stacked cell architecture search}
\label{sec:searching}
\subsubsection{Cell structure design}
Traditional NAS methods mainly search over a discrete set of candidate architectures, which are very computationally intensive. The recent work \citep{liu2018darts} proposed to solve the problem in a differentiable way. The computing architecture (i.e., cell) is represented as a directed acyclic graph. As shown in Fig.~\ref{fig:cells}, an ordered sequence of N nodes ${x_1,\dots,x_N}$ (latent representation, e.g. feature map) are connected by directed edges $(i,j)$ (i.e. operations $o(\cdot) \in O$). The NAS process boils down to find the optimal cell structure. 

\begin{figure}[htbp]
	%  \vspace{-3mm}
	\centering
	\includegraphics[width=0.9\linewidth]{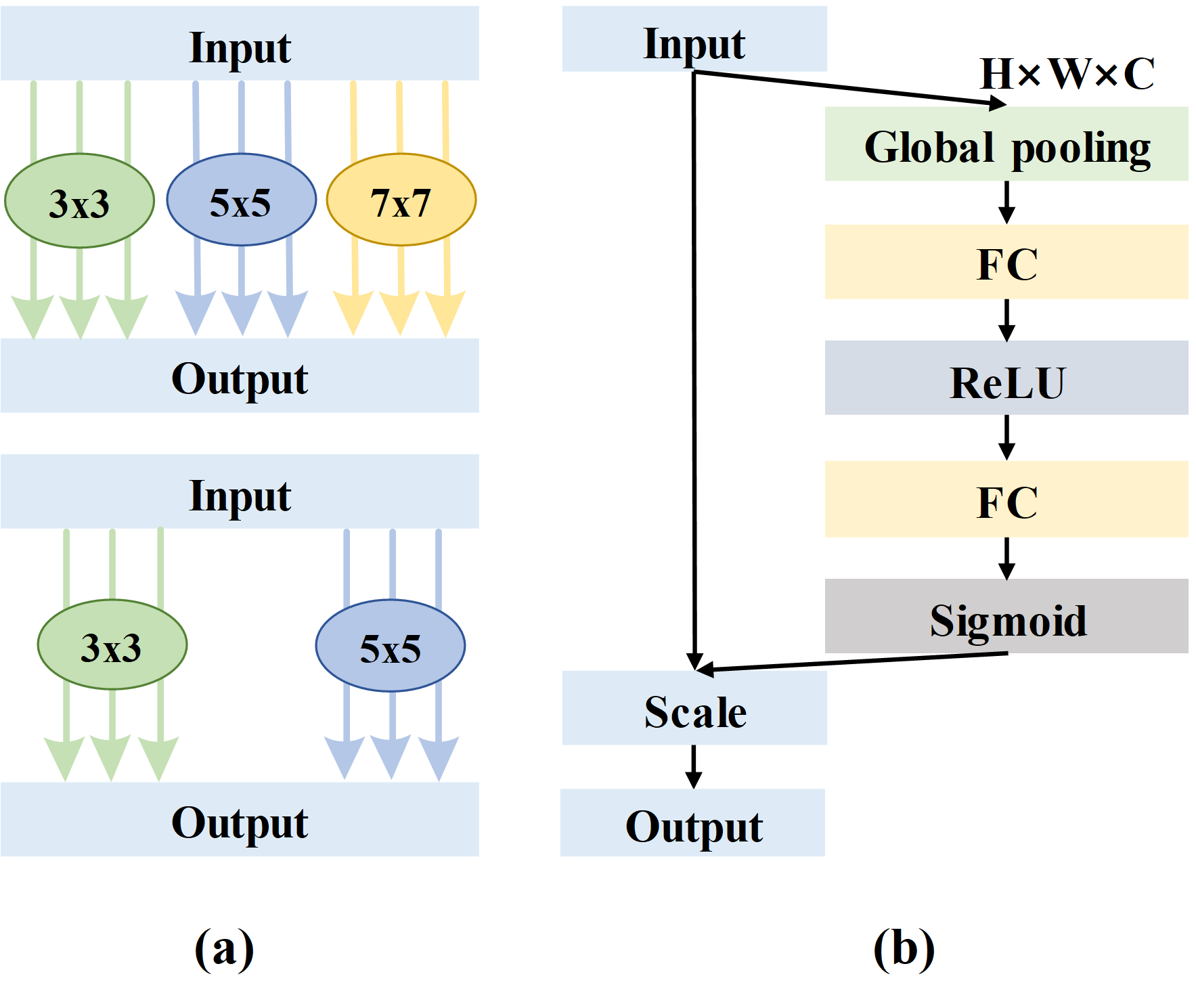}
	%   \vspace{-3mm}
	\caption{The detail of MixConv (a) and SE block structure (b).}
	
	\vspace{-1mm}
	\label{fig:mixconv}
\end{figure}

Formally, each internal node is computed as the sum of all its predecessors within a cell:
% {\setlength\abovedisplayskip{1pt plus 3pt minus 2pt}
%  \setlength\belowdisplayskip{1pt plus 3pt minus 2pt}
\begin{equation}\label{equ:searchspace}
x_j=\sum_{i\leq j}{O^{(i,j)}(x_i)}
\end{equation}

In each cell, there are candidate operations between two different nodes, including none, identity, max-pooling, separable convolution, and dilation convolution. To further compress the model and improve the performance, we introduce the MixConv layer and SE block (see Fig.~\ref{fig:mixconv}) into the candidate operations.

For example, a MixConv35 layer combines the convolutional kernels of sizes $[3\times3]$ and $[5\times5]$ to fuse the features extracted from different receptive fields. SE block exploits adaptive channel-wise feature recalibration to boost feature learning and model generalization ability. After the searching process, the two candidate operations with the highest probability value are selected for each edge.

\begin{figure}[htbp]
	\centering 
	\subfloat{
		\label{cell:normal}
		\includegraphics[width=1.0\linewidth]{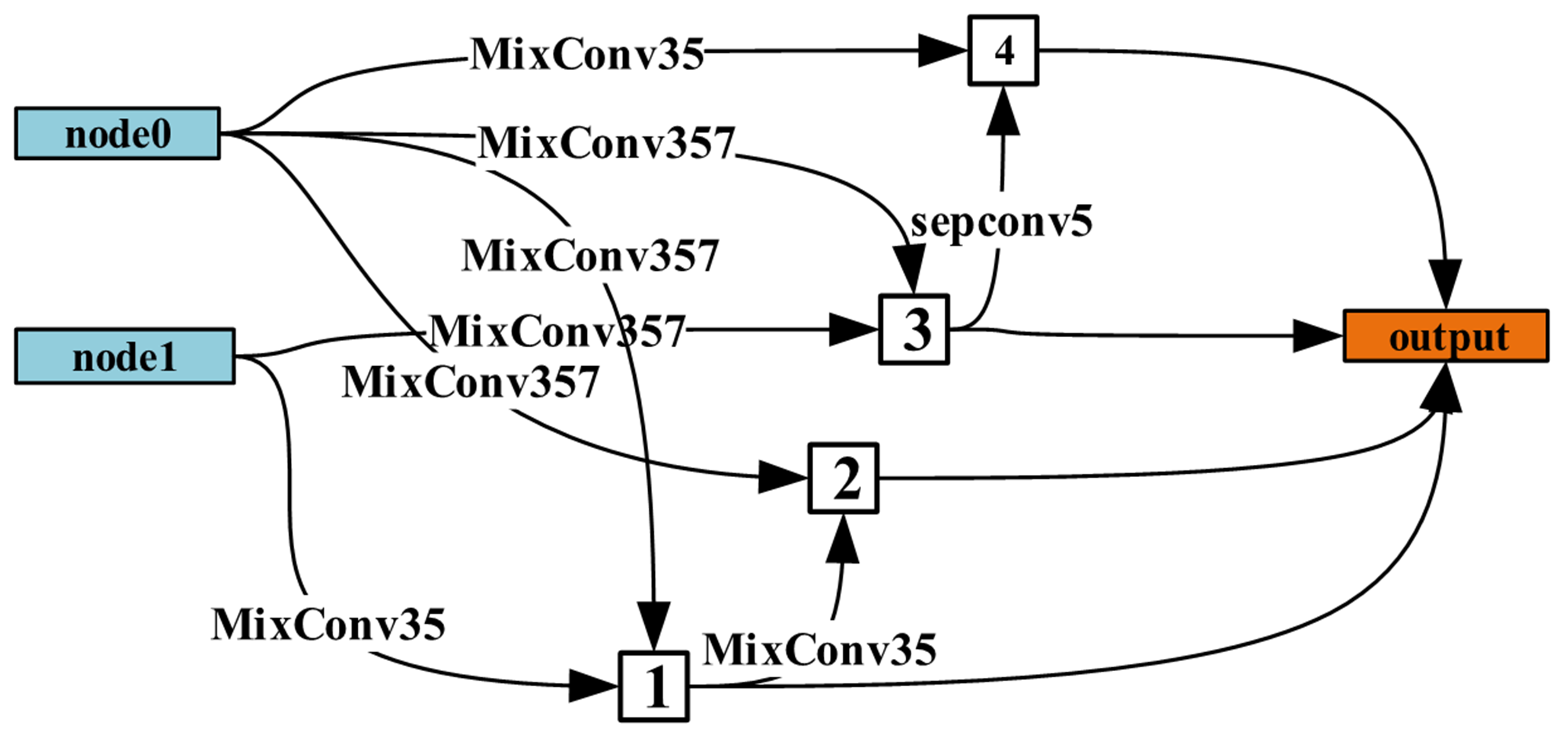}
	}
	\quad
	\subfloat{
		\label{cell:reduction}
		\includegraphics[width=1.0\linewidth]{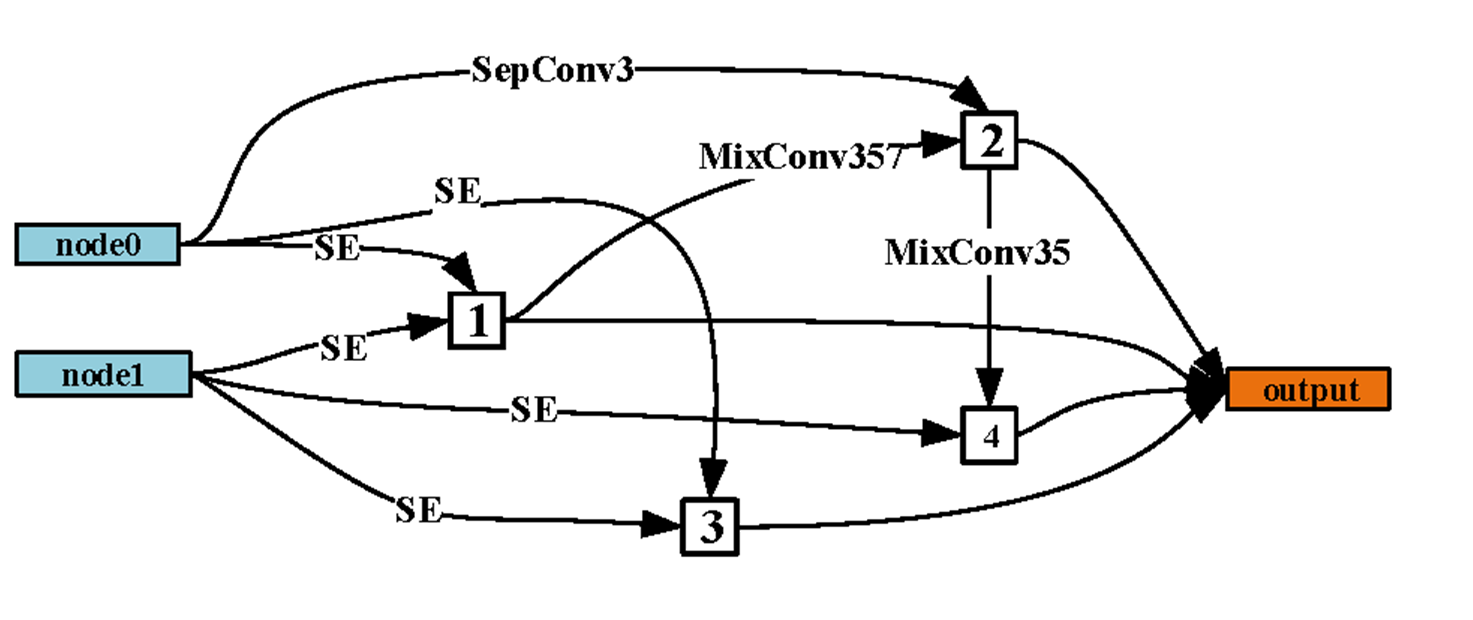}
	}
	\caption{The configuration of the searched cells in the classification (top) and segmentation (bottom) networks. There are 7 nodes in each cell: 2 input nodes (node0, node1), 4 internal nodes (boxes 1, 2, 3, 4), and 1 output node. The input nodes take the outputs of two previous cells as input, and connect with the internal nodes through the candidate operations. Then, the results of internal nodes are concatenated to obtain the output of this cell.}
	\label{SearchedCell}
\end{figure}

Fig.~\ref{SearchedCell} presents the configuration of a classification cell and a segmentation cell obtained after the searching process. They both share the same candidate operations. For different tasks, the searched cell structures have significant divergence. It can be observed that the classification cells are more likely to use MixConv as well as larger kernels, since learning scale-invariant features is more important for the classification task. In contrast, the segmentation cell prefers the SE operation. The recent work \citep{rundo2019use} indicates that the SE block’s adaptive feature recalibration provides excellent cross-dataset generalization, thus boosting the segmentation performance. Considering our dataset is collected from different equipment, such searched results are reasonable.

\subsubsection{Progressive growing strategy}
The parameters in our framework consist of the network parameters $\alpha$ and weight of architecture $\omega$. They are simultaneously learned through a bi-level optimization strategy:
\begin{equation}\label{bi-opti}
\begin{split}
\min\limits_\alpha \quad  & L_{val}(\omega^{\ast}(\alpha),\alpha),\\
s.t. \quad & \omega^{\ast}(\alpha)=\arg \min \limits_\omega L_{train}(\omega,\alpha).\\
\end{split}
\end{equation}
where $L_{train}$ is the training loss and $L_{val}$ is the validation loss. We optimize the weight of the architecture $\omega$ in the inner level and the network parameters $\alpha$ in the outer level respectively.
The final architecture is then derived by:
\begin{equation}\label{derive}
o^{(i,j)}=\mathop{\arg\max}\limits_{o\in O,o\neq {zero}}\frac{\exp(\alpha^{(i,j)}_o)}{\sum_{o^\prime \in O}{\exp(\alpha^{(i,j)}_{o^\prime})}}
\end{equation}
Where $o^{(i,j)}$ denotes the operation of transformation from $node^i$ to $node^j$. $O =\{o^{(i,j)}\}$ is the set of candidate operations. It can be seen from Eq. \ref{bi-opti} that the optimization of the network parameters $\alpha$ and the weight of architecture $\omega$ is intertwined and requires a good balance. 

\begin{figure}[htb]
	%  \vspace{-3mm}
	\centering
	\includegraphics[width=0.98\linewidth]{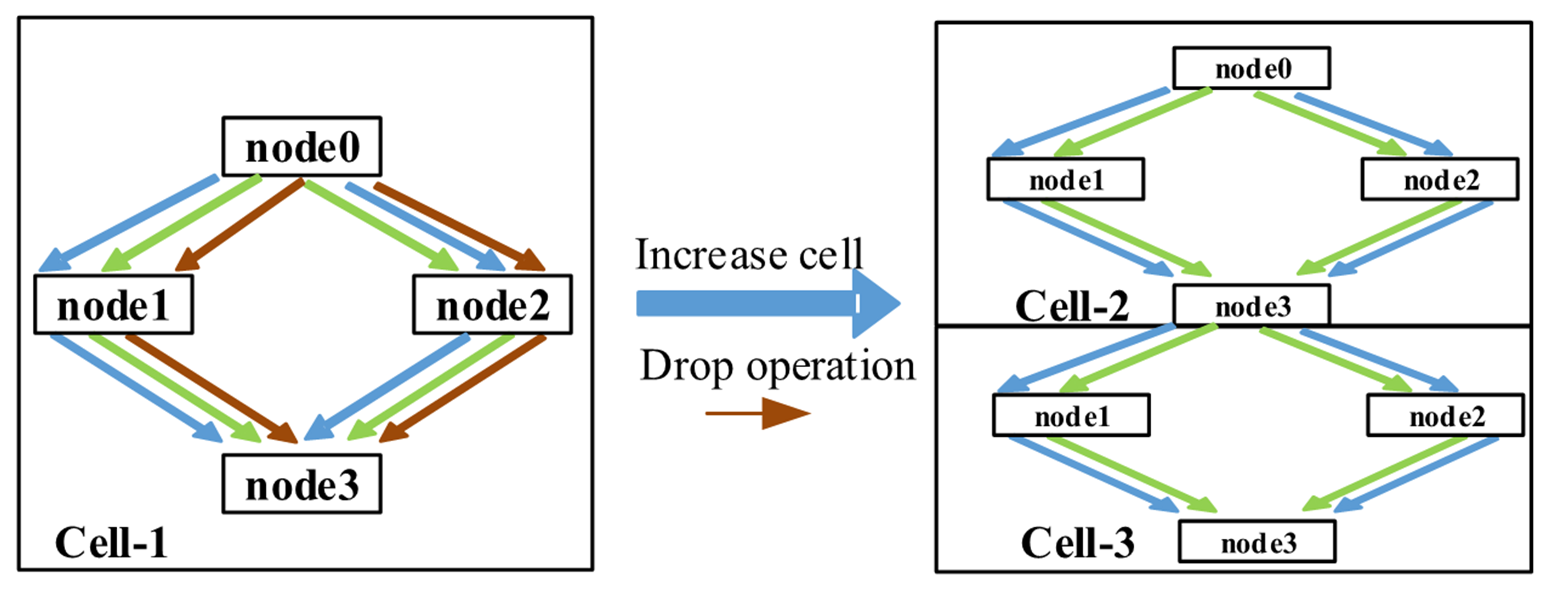}
	\caption{The progressive search process. The number of cells is increased while one of the operations was dropped during growth (red arrows in Cell-1).}
	\label{fig:cells}
	% \vspace{-5mm}
\end{figure}

As illustrated in Fig.~\ref{fig:cells}, we propose a progressive growing strategy to address this problem. As the searching process going on, the number of the stacked cells increases hierarchically while the candidate operations are dropped one-by-one. In other words,  there are a lot of operations in the early stage of the search, and we need to focus on the optimization of the architecture $\omega$. Then, we switch the computation cost from the architecture to the network parameters $\alpha$ by reducing the candidate operations and constructing a deeper network. Thus, the search process can focus more on the inner optimization of the network in the deep stages. Additionally, modern neural networks are usually very deep (e.g. ResNet152\citep{he2016deep}). Our proposed searching strategy is able to progressively build a deep network architecture, which is crucial for extracting discriminative features.

Specifically, the cell architecture searching is divided into $K$ stages. In each stage, we stack a new cell behind an old one (see Fig.~\ref{fig:cells}). The newly stacked cell has the same network parameters as the old one. The operation that obtains the lowest score (i.e. highest validation loss) will be dropped stage-by-stage. The progressive growing process can be formulated as follows:
\begin{equation}\label{progressive}
\begin{split}
o^{(i,j)}_k &\rightarrow o^{(i+1,j)}_{k+1}(o^{(i,j+1)}_{k+1})\\
\|O_{k+1}\| &= \|O_{k}\|-1
\end{split}
\end{equation}
Where $k=1,2,\cdots, K$ indicates the stage index, and $\|O\|$ denotes the cardinal number of candidate operations set.

\subsection{Cell re-aggregation}
To reduce the memory cost and search time during training, cells are often stacked in a series-connected mode\citep{liu2018darts,liu2018progressive}. However, such structure mode limits the feature extraction ability of the network. It also has an adverse effect on the network capturing context information due to the unsuitable receptive field size. Therefore, we introduce the expert-designed structure to re-aggregate those stacked cells after the search completes. Our proposed re-aggregation significantly improves the model performance for the echinococcosis classification and ovary segmentation. 

\begin{figure}[t]
	\centering
	\includegraphics[width=1.0\linewidth]{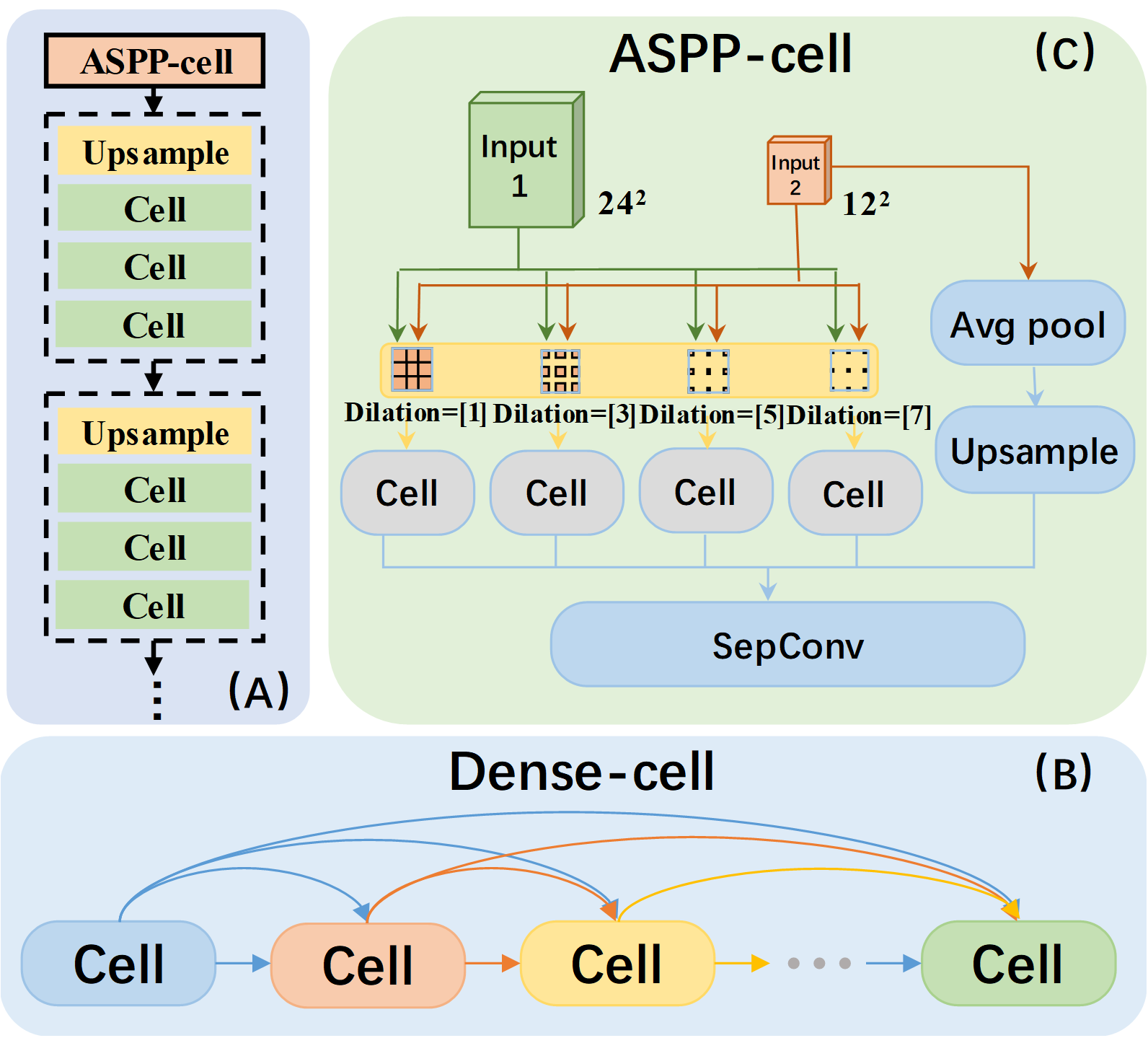}
	\caption{The details of segmentation decoder flowchart (A), Dense-cell structure (B), and ASPP-cell structure (C)}
	\label{fig:cell_aggregation}
\end{figure}

As shown in Fig.~\ref{fig:cell_aggregation}, we regroup the stacked cells into a dense structure (dense-cell) and an atrous spatial pyramid pooling structure (ASPP-cell) for the classification and segmentation tasks respectively. In the dense-cell structure (Fig.~\ref{fig:cell_aggregation}B), we introduce direct connections from any cell to all subsequent cells. The input information will be more sufficient for each cell while the operation of each cell remains unchanged as in the search time.

The ASPP-cell is applied at the beginning of the decoder cell block (Fig.~\ref{fig:cell_aggregation}A). Specifically, as shown in Fig.~\ref{fig:cell_aggregation}C, the ASPP-cell receives two inputs with different sizes (i.e. $24\times24$ and $12\times12$). Four parallel atrous convolutions with different atrous rates are applied to the inputs. Then, the searched cells follow behind to process the feature maps. To alleviate the problem of filter weights malfunction and incorporate the global information, we apply average pooling at the lower-resolution input and upsample it by a factor of 2. Finally, the separable convolution is applied to the aggregated feature maps and generates the output. 

We believe that introducing human prior knowledge of designing a neural network is beneficial to model adaptability in different tasks. The Dense-cell structure reformulates the information flow in the architecture which strengthens the recognition ability in classification tasks. The ASPP-cell structure effectively captures multi-scale information which acts as a significant role in segmentation tasks. These structures are carefully designed and their effectiveness has been validated\citep{huang2017multi,chen2017rethinking,li2020deep}. Our cell re-aggregation introduces these structures to augment collaboration among cells, which contributes to obtaining a high-performance model. Additionally, different tasks have specialized prior preferences. It is easy to generalize our cell re-aggregation scheme in other circumstances.

\section{Experiments and results}
In this section, we first describe the datasets, implementation details, and evaluation metrics. We then report the results of echinococcosis classification and ovary segmentation. Comparison with other methods and ablation studies are also performed.

\subsection{Datasets and pre-processing}
To comprehensively evaluate our proposed HASA framework, we performed experiments on two large US image datasets: 1) an echinococcosis dataset for the classification task; 2) an ovary dataset for the segmentation task. For both tasks, our method achieved the best performance with less computational cost. This study was approved by the local institutional review board. Informed consent was waived because only deidentified data was used. \par

\subsubsection{Classification dataset}
We first tested our method for the classification of hepatic echinococcosis. Echinococcosis is a zoonotic parasitosis caused by a larval infestation that mainly occur in the liver \citep{mcmanus2003echinococcosis}. The two main types of the disease are cystic echinococcosis (CE) and alveolar echinococcosis (AE). According to disease progression, CE can be classified into CL, CE1, CE2, CE3, CE4, and CE5, and AE is classified into AE1, AE2, and AE3 (see examples for different classes in Fig.~\ref{example_pic}). AE is caused by infection with echinococcus multilocularis and starts with infiltration type (AE1) which is characterized by a tumor-like, infiltrative, and destructive growth. AE2 is characterized by the occurrence of calcium salt deposition and calcification. Finally, the lesion proliferates into a huge lesion, and the central part of the lesion becomes hollow due to ischemic necrosis and liquefactive necrosis, which forms the liquefaction necrosis type (AE3). Details about the classification of CE can be found here \citep{who2003international}. 

US is the preferred imaging modality for echinococcosis diagnosis \citep{mcmanus2003echinococcosis}; however, the diagnosis is non-trivial due to the complex sonographic appearance and high intra- and low inter-class variety (Fig.~\ref{example_pic}). It is a tough task which requires the strong discriminative power from the network design. The classification dataset consists of 9566 US images from 5028 patients who were diagnosed with hepatic echinococcosis. B-mode and convex probes were used. The image numbers for each class are as follows: AE1 673, AE2 261, AE3 183, CL 2050, CE1 1595, CE2 399, CE3 1499, CE4 1974, and CE5 932. Some patients had more than one lesion in the liver, and for some lesions more than one image was saved by the doctors. Due to these reasons, the image number is not the same as the patient number. The ground-truth class label of each lesion was provided by abdominal sonographers with at least 5-year experience and further reviewed by a senior expert with 21-year experience. The lesions were manually annotated by sonographers with bounding boxes. All lesions were cropped according to the bounding boxes and resized to 224$\times$224. The dataset was then divided into the training and test sets according to a ratio of 7$:$3. \par

\subsubsection{Segmentation dataset}
We further tested our method on an ovary dataset for ovary and follicle segmentation. An ovary image contains multiple follicles. Accurately segmenting the ovary and follicles is non-trivial \citep{seg2}. The main challenges for automatic segmentation include very large variations in the size of ovary and follicles, poor contrast of ovary against the background, and touching boundary among follicles. The ovary ultrasound images were obtained by transvaginal ultrasonography. Ten sonographers with at least 5-year experience manually delineated all the ovaries and follicles in each image, and the results were reviewed by a senior expert with 20-year experience in pelvic ultrasound. The dataset consists of 3204 images, in which 2509 images from 153 patients were selected as the training set. The rest were used as the test set. All images were resized to 448$\times$448.

To avoid potential bias, when splitting data we make sure that the images from the same patient went into either the training or test set. In addition, data augmentation strategies including normalization, rotation ($\pm$20$^{\circ}$), and random horizontal flip were applied to the training images. 

\begin{table*}[htbp]
	\centering
	% \vspace{-8mm}
	\caption{Quantitative comparisons of different models on classification task}
	\begin{tabular}{c|ccccccc} 
		\toprule
		Method  &Pre-trained &Param(M)  &Size(MB) &Accuracy($\%$) &Precision($\%$) & Recall($\%$) &F1-score($\%$)   \\\midrule
		DarkNet53       &yes    &41.61  &333.2  &85.71 &85.76 &85.71   &85.65  \\ \hline
		DarkNet53       &no     &41.61  &333.2  &83.33 &83.47 &83.33   &83.27  \\ \hline
		SPOS            &no     &4.5    &26.98  &83.04 &84.03 &83.04   &82.76  \\ \hline
		ProxylessNAS    &no     &5.41   &41.5  &83.71 &84.15 &83.71   &83.61  \\ \hline
		DARTS   &no &4.35           &  34.1     &  83.75     & 83.83      &83.75        & 83.54    \\ \hline
		P-DARTS &yes  &4.18          & 33.8    &  84.62     &84.69       &84.62       &84.60           \\ \hline	
		P-DARTS        &no&   4.23        &  30.5     &     84.08   & 84.21      & 84.08      &84.07    \\ \hline
        Pruning &yes &30.56 &228.31 &84.29 &84.70 &84.29 &84.49 \\ \hline
		
		HAS-stage5  &yes &26.30    & 201    &  85.50     &85.77       &85.50       &85.53    \\ \hline
		
		HAS-w/o-MS &yes &27.39    &218  &86.08    & 86.18  & 86.08 & 86.04    \\ \hline
		
		\textbf{HAS} &yes &26.59  &204 &86.67 &86.73 &86.67 &86.64    \\ \hline
		
		\textbf{HASA$_{Dc}$} &yes &26.59  &204 & \textbf{87.83} & \textbf{88.05}  &\textbf{87.83}  &\textbf{87.86}    \\ 
		
		\bottomrule
	\end{tabular}
	\label{tab:class_Table}
	% \vspace{-5mm}
\end{table*}

\subsection{Implementation details}
In this paper, all the experiments were carried out with PyTorch. Our framework was trained on a single Titan X GPU with 12GB RAM (Nvidia Corp, Santa Clara, CA). Specific experimental setting for the image classification and segmentation tasks are described as follows.

\subsubsection{Classification setting}
The pre-trained DarkNet53 was used as the classification backbone. DarkNet53 is the backbone structure of the famous detection model YOLO\citep{redmon2016yolo}.  Inheriting the strengths of ResNet, DarkNet53 has excellent representation learning ability and  can avoid the vanishing gradient problem when training a very deep neural  network. The progressive growing strategy was used for stacking the cell architectures. Specifically, there were five initial cells at the beginning. In each stage, we stacked two cells behind. K was set to three. As a result, there were a total of 11 cells involved in the NAS search. The architecture search model was trained for 25 epochs in each stage.  Cross entropy loss was used for the classification task. We set the batch size to 36 and used Adam optimizer with a learning rate of 0.025 to update the network parameters. The search time for finding our final classification architecture is about 1.5 hours, and the time for training the whole classification network is about 2.5 hours. \par

\subsubsection{Segmentation setting}
We adopted the pre-trained MixNet as the segmentation backbone. MixNet naturally integrates the dimensions of multiple convolution cores into a single convolution, which reduces the amount of convolution computation while potentially improving the accuracy and efficiency of the model. Several cell structures were stacked behind as a block, including a cell and bilinear interpolation at the beginning. There were a total of two blocks involved in the progressive growing strategy. In each stage, we stacked one cell in each block. The K was set to two. As a result, there were a total of six cells involved in the NAS search. Finally, we used a 4$\times$ up-sampling to generate the prediction directly, with the same resolution as the input. It is worth noting that the candidate operations do not contain max-pooling and average-pooling. Dice similarity coefficient (DSC) loss was used for the segmentation task. We set the batch size to 8 and used Adam optimizer with a learning rate of 0.0001 to optimize the model. The search time for finding our final segmentation architecture is about 9.5 hours, and the time for training the whole segmentation network is about 17.5 hours. \par

\subsection{Evaluation metrics}
In this paper, we focus on the evaluation of the echinococcosis classification accuracy and the ovarian segmentation accuracy of the HASA framework. The parameter size is also taken into account as a significant evaluation metric in the experiments. For the echinococcosis classification, we used four metrics to evaluate model performance, including accuracy, precision, recall, and F1-score. For the ovary segmentation, we also used four metrics, including DSC, Jaccard coefficient (JC), Hausdorff distance (HD), and average surface distance (ASD). DSC and JC are commonly employed to evaluate the region similarity between segmentation and annotation. HD and ASD denote the longest and average distance over the shortest pixel distances between the surface of segmentation and annotation. Lower HD and ASD, and higher DSC and JC denote better segmentation performance.

\subsection{Echinococcosis US classification performance}
We first tested our method in classifying echinococcosis into nine subtypes in US images. We compared our method with other methods and performed ablation studies to prove the effectiveness of the components of our method. We then showed the confusion matrix and t-SNE visualization. Finally, we presented a few example images and the predicted labels by different methods. 

\subsubsection{Comparison with other methods}
To verify the effectiveness of the proposed method for echinococcosis classification, we compared it with several other DNNs, including:

\textbf{DarkNet53}: DarkNet53 \citep{redmon2018yolov3} serves as a baseline for comparison since our method replaces the 4$^{th}$ and 5$^{th}$ stages of DarkNet53 with the searched cells. DarkNet53 mainly consists of convolutional layers and residual structures, which have powerful feature learning capabilities.

\textbf{SPOS}: SPOS \citep{guo2019single} is an evolution learning based search method. it constructs a simplified supernet, where all architectures are single paths so that weight co-adaption problem is alleviated. All architectures are trained equally with a uniform path sampling strategy.

\textbf{ProxylessNAS}: Unlike proxy NAS algorithms that search for building blocks on proxy tasks such as training for fewer epoch or stating with a smaller dataset, proxylessNAS \citep{cai2018proxylessnas} directly learns the architectures on the target task without any proxy and significantly reduces the computational cost of architecture search.

\textbf{DARTS}: Instead of searching for architectures over a discrete and non-differentiable search space, DARTS \citep{liu2018darts} relaxes the search space to be continuous, thus allowing efficient search of the architecture using gradient descent.

\textbf{P-DARTS}: Due to the large gap between the depths in search and evaluation scenarios, previous differentiable search methods suffer from lower accuracy in evaluating the searched architecture or transferring it to another dataset. P-DARTS \citep{chen2019progressive} uses search space approximation and regularization to allow the depth of searched architectures to grow gradually during training.

\textbf{Pruning}: To test whether pruning approaches can increase the performance of DarkNet53 with a smaller number of parameters, we adopted the L1-norm based channel pruning approach \citep{li2016pruning}. For a fair comparison, we balanced the pruning process to ensure the pruned model had the same-level parameters and size as ours.

For a fair comparison, all NAS-based experiments used a similar training scheme. Specifically, the training data is further separated into a 30\% subset for architectures search and the remaining 70\% for updating the network parameters. Table~\ref{tab:class_Table} presents the accuracy, precision, recall, and F1-score of different models. Compared with the DarkNet53, the proposed HASA$_{Dc}$ achieved better performance with less resource consumption. Specifically, our HASA$_{Dc}$ reduced about 40$\%$ parameters compared with DarkNet53 while achieving $2\%$ higher F1-score. The pruning approach significantly reduces the parameters but slightly degrades the F1-score by 1.2 percent compared with the pre-trained DarkNet53 without pruning. The HASA$_{Dc}$ also got much better performance compared with the models searching from scratch (i.e., SPOS, proxylessNAS, DARTS, and P-DARTS). These results proves that the proposed HAS framework is superior to the vanilla NAS models in the classification task. We conjecture that this is due to the better feature extraction ability of our hybrid model. \par

By default our HAS replaces the 4$^{th}$ and 5$^{th}$ stages with stacked cells. We also implemented several variants of our method to compare different architectures of the hybrid model:

\textbf{HAS-stage5}: We only replaced the 5$^{th}$ stage with stacked cells. The channels of the cells were set to $512,1024$ for being consistent with DarkNet53.

\textbf{HAS-w/o-MS}: To demonstrate the advantage of introducing the MixConv and SE block into the candidate operations, we experimented with the HAS framework without the MixConv and SE block operations.

\textbf{HASA$_{Dc}$}: We added the Dense-cell to our HAS framework to show the effectiveness of this component

In Table~\ref{tab:class_Table}, it is interesting to note that HAS performed better than HAS-stage5. This indicates that only replacing the layers close to the final classification layer is not enough. Meanwhile, the HAS had higher accuracy than HAS-w/o-MS, which is attributed to the introduction of the MixConv and SE block operations. After introducing the Dense-cell, the resulting HASA$_{Dc}$ further improved the classification performance on all the metrics, which shows the efficacy of our re-aggregation strategy. \par 

We then performed  paired-sample t-test to see whether the difference of performance is significant between our HASA$_{Dc}$ and other methods. Specifically, the probabilities of belonging to the ground-truth class predicted by our method and the other method were used as the scores for each sample. P-value less than 0.05 is considered statistically significant. As shown in Table~\ref{tab:class_Ttest}, our proposed HASA$_{Dc}$ performs statistically better than all other methods. \par

\begin{table}[htbp]
	\centering
	% \vspace{-8mm}
	\caption{Significance test on classification task}
	\begin{tabular}{c|cc} 
		\toprule
		Method  &Pre-trained &P-value   \\\hline
		DarkNet53       &yes    &1.1$\times 10^{-4}$  \\ \hline
		DarkNet53       &no     &1.1$\times 10^{-5}$  \\ \hline
		SPOS            &no     &4.4$\times 10^{-12}$    \\ \hline
		ProxylessNAS    &no     &5.0$\times 10^{-9}$ \\ \hline
		DARTS           &no     &9.4$\times 10^{-6}$      \\ \hline
		P-DARTS         &yes    &4.3$\times 10^{-7}$       \\ \hline	
		P-DARTS         &no     &9.5$\times 10^{-11}$   \\ \hline
		
		HAS-stage5      &yes    &1.6$\times 10^{-4}$  \\ \hline
		
		HAS-w/o-MS      &yes    &1.3$\times 10^{-3}$  \\ \hline
		
		HAS             &yes    &5.6$\times 10^{-3}$  \\ 
		\bottomrule
	\end{tabular}
	\label{tab:class_Ttest}
	% \vspace{-5mm}
\end{table}

Moreover, we evaluated the effect of the number of searched cells used in the evaluation stage on the classification accuracy of HASA$_{Dc}$. Note that the number of cells used in the evaluation stage does not have to be the same as that in the search stage. As in previous studies on NAS \citep{chen2019progressive, liu2018darts}, the purpose of the search stage is to discover the best cell architecture. It can be observed in Table~\ref{tab:cell_num} that as the number of cells increased from 3 to 7, the accuracy of the model improved by 2.33\%. However, further increasing the cell number did not improve the performance but required higher computational costs. This is consistent with the general knowledge that to some extent the performance of the model improves as the number of parameters increases. Therefore, we need to balance the size and performance of the model.

\begin{table}[htbp]
	\centering
	\caption{The classification accuracy of HASA$_{Dc}$ with different number of searched cells in the evaluation stage. }
	\begin{tabular}{c|cc}
		\toprule
		\textbf{Cell Number} &\textbf{Size(MB)} &\textbf{Accuracy($\%$)} \\
		\hline
		3 &134 &85.50 \\ 
		4 &145 &86.00 \\
		5 &161 &86.08 \\
		6 &180 &87.17 \\
		\textbf{7} &\textbf{204} &\textbf{87.83} \\
		8 &231 &86.58 \\
		9 &263 &85.83 \\
		\bottomrule
	\end{tabular}
	\label{tab:cell_num}
\end{table}

\subsubsection{Confusion matrix and t-SNE visualization}
\begin{figure}[htbp]
	%  \vspace{-3mm}
	\centering
	\includegraphics[width=1.0\linewidth]{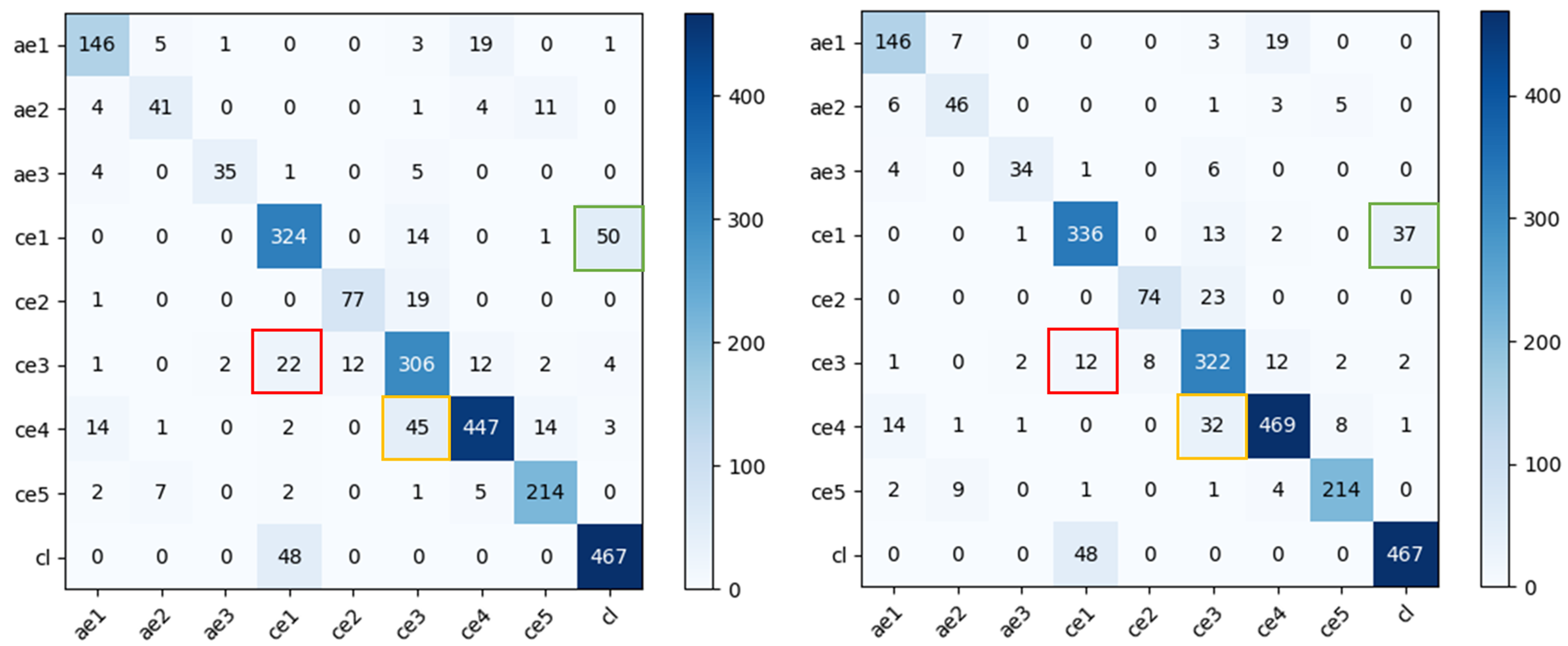}
	%   \vspace{-3mm}
	\caption{The confusion matrix of the DarkNet53 (left) and the proposed HASA$_{Dc}$ (right). The vertical axis is the ground truth.}
	\vspace{-1mm}
	\label{fig:confusion}
\end{figure}

Fig.~\ref{fig:confusion} shows the confusion matrices of the baseline DarkNet53 and the proposed HASA$_{Dc}$. It can be seen that the classification performance of HASA$_{Dc}$ was improved compared with that of DarkNet53. Specifically, most off-diagonal values of HASA$_{Dc}$ were smaller than those of DarkNet53 (examples are shown in the red, yellow, and green boxes), which indicates that fewer samples were misclassified by HASA$_{Dc}$.

\begin{figure}[t!]
% 	\vspace{1.14mm}
	\centering
	\includegraphics[width=0.8\linewidth]{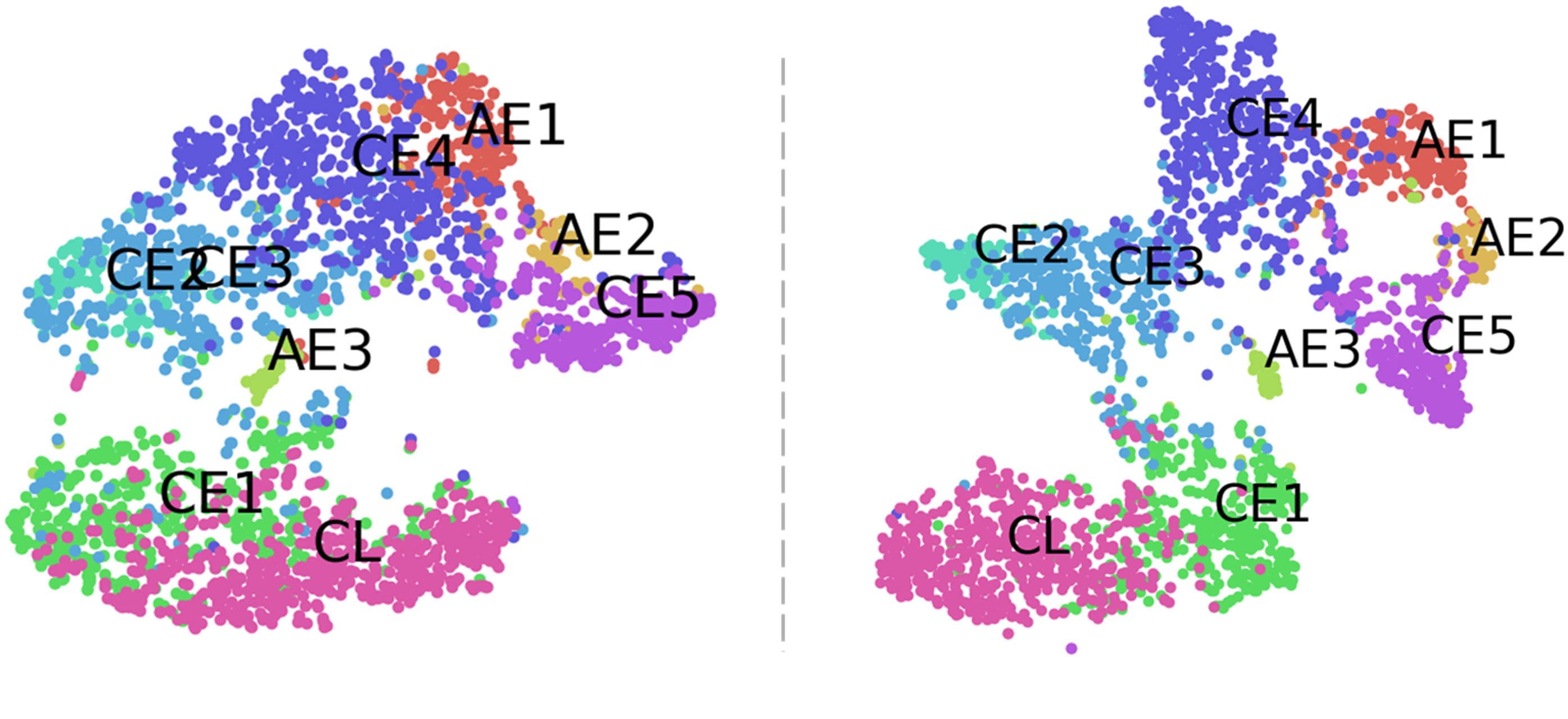}
	%   \vspace{-3mm}
	\caption{The t-SNE visualization of the DarkNet53 (left) and our proposed HASA$_{Dc}$ (right).}
	\vspace{-1mm}
	\label{fig:tsne}
\end{figure}

To visualize the learned features by the two models, we used t-SNE to display the 2D embedding. Fig.~\ref{fig:tsne} (left) is the embedding learned by the vanilla DarkNet53 network. It corresponds well with the clinical experience of doctors, where CL and CE1, CE4 and AE1, CE5 and AE2 are difficult to be differentiated as they are similar to each other. Fig.~\ref{fig:tsne} (right) shows the embedding learned by the HASA$_{Dc}$. Note that the samples of the same class became closer and those challenging classes became more separable. This proves that the proposed HASA$_{Dc}$ is more capable of disentangling the inter-class similarities, which is vital to the fine-grained classification.

\subsubsection{Examples of image classification}
\begin{figure}[htbp]
	\centering
	\includegraphics[width=1.0\linewidth]{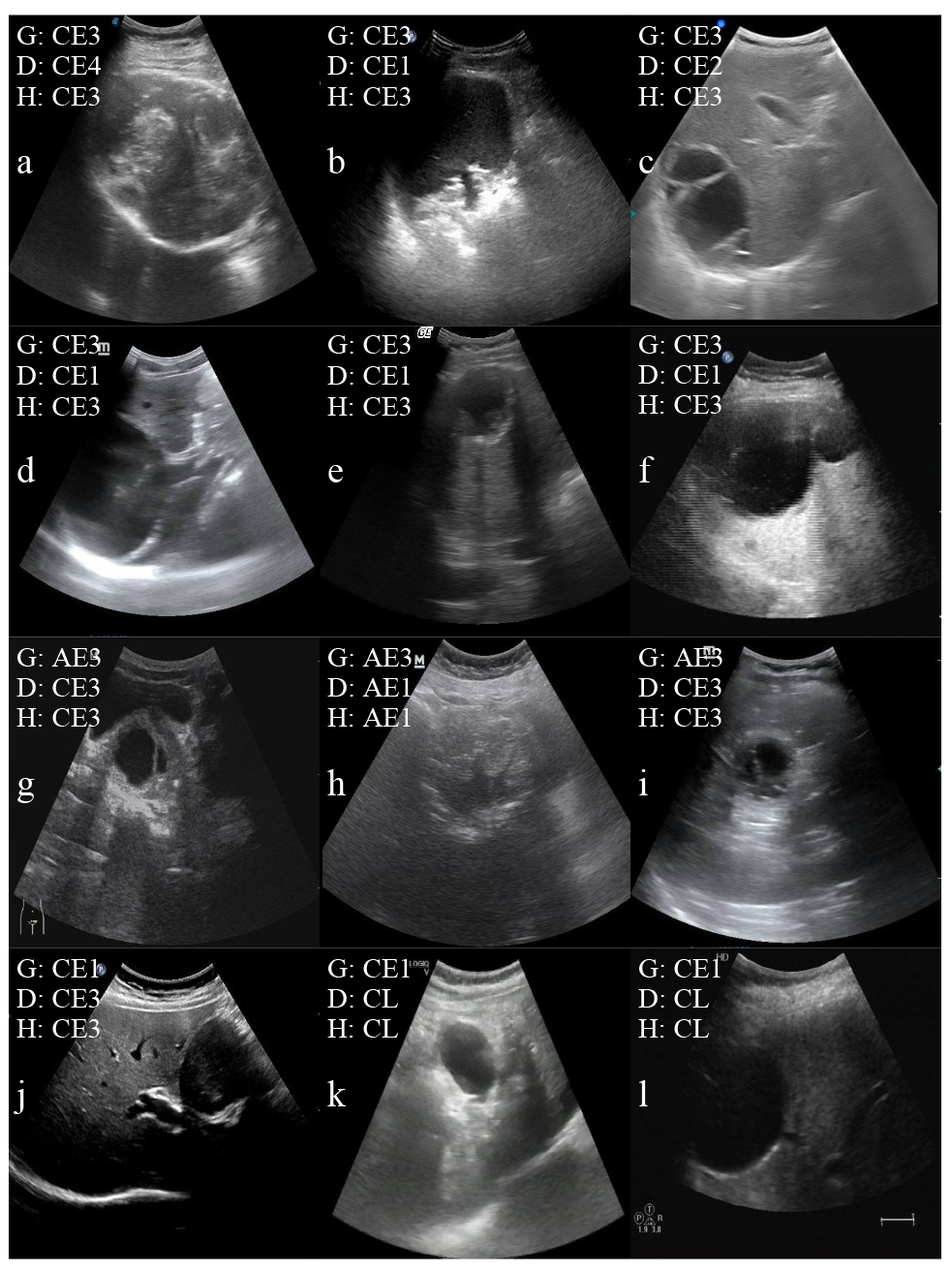}
	\caption{Examples of the image classification results of the pre-trained DarkNet53 and our method HASA$_{Dc}$. The ground-truth label (G) and the predicted segmentation by the DarkNet53 (D) and HASA$_{Dc}$ (H) are shown in the top left corner.}
	\label{fig:classificationDemo}
\end{figure}

We also show some example images along with the predicted labels by the pre-trained DarkNet53 and our method HASA$_{Dc}$ in Fig.~\ref{fig:classificationDemo}.  The images in the first two rows all belong to CE3. These images were misclassified into CE1, CE2, or CE4 by the DarkNet53, which is consistent with the statistics in the confusion matrix in Fig.~\ref{fig:confusion}. In contrast, our HASA$_{Dc}$ successfully classified them. There are also cases that both methods failed. In the third row, both DarkNet53 and HASA$_{Dc}$ misclassified AE3 into AE1 or CE3. The misclassification of AE3 into AE1 (Fig.~\ref{fig:classificationDemo}h) is probably due to the poor contrast of the lesion. The last row shows three examples of CE1 that were misclassified into CE3 or CL by both methods. As shown in the confusion matrix in Fig.~\ref{fig:confusion}, CL and CE1 are the most difficult classes to be differentiated. They are both unilocular, cystic lesions with uniform anechoic content. Their difference is that for CE1 the cyst wall is visible while for CL it is not. However, the two methods may not well capture this very subtle difference. \par

\subsection{Ovarian US segmentation performance}
We then tested our method in another task: segmenting the ovary and follicles in US images. We compared the segmentation performance of our method and other methods and showed a few examples of the segmentation results by different methods.

\subsubsection{Comparison with other methods}
We compared our method with the the following methods to verify its effectiveness in the segmentation task: 
 
\textbf{DeepLabv3}: DeepLabv3 \citep{chen2017rethinking} addresses the problem of segmenting objects at multiple scales. It consists of modules that use atrous convolution in cascade or in parallel to capture multi-scale context by adopting multiple atrous rates. In the original DeepLabv3, Xception  was used as the backbone. In our study, since we used MixNet as the backbone in our method, we also tested DeepLabv3 with the MixNet as the backbone and referred to this setting as the baseline.

\textbf{U-Net}: Based on fully convolutional network, U-Net \citep{ronneberger2015u} modifies and extends this architecture to enable training with fewer images and yielding more precise segmentation. It consists of a contracting path to capture context and a symmetric expanding path to enable precise localization.

\textbf{Gated-SCNN}: Instead of processing color, shape, and texture information all together in a deep network, Gated-SCNN \citep{takikawa2019gated} is a two-stream architecture for semantic segmentation that explicitly wires shape information as a separate processing branch that processes information in parallel to the classical stream. Higher-level activations in the classical stream are used to gate the lower-level activations in the shape stream, which can remove noise and let the shape stream focus on processing the boundary-related information.

\textbf{Pruning}: To test whether the segmentation performance of the baseline (i.e., DeepLabv3 using MixNet as backbone) can be improved with a smaller number of parameters using the L1-norm based channel pruning approach \citep{li2016pruning}, we controlled the pruning process to ensure the pruned model had the same-level parameters and size as our method for a fair comparison.

\textbf{SPOS, ProxylessNAS, and DARTS}: Similar to the proposed hybrid framework, the searched architectures by the three NAS algorithms \citep{guo2019single,cai2018proxylessnas,liu2018darts} were respectively combined with the MixNet backbone.

\textbf{HASA$_{Ac}$}: HASA$_{Ac}$ represents our HAS framework with ASPP-cell. This is to verify the effectiveness of the cell re-aggregation strategy.

\begin{table*}[htb]
	\centering
	\caption{Quantitative comparisons of different methods on segmentation task}
	\resizebox{\textwidth}{15mm}{
	\begin{tabular}{c|c|cccc|cccc|c}
		\hline
		\multirow{2}*{\textbf{Method}}  &\multirow{2}{*}{\textbf{Backbone}}
		&\multicolumn{4}{c|}{\textbf{Follicle}} &\multicolumn{4}{c|}{\textbf{Ovary}}  &\multirow{2}*{\textbf{Param(M)}}\\
		\cline{3-10}
		& & DSC(\%)$\uparrow$ &JC(\%)$\uparrow$ &HD(pix)$\downarrow$ &ASD(pix)$\downarrow$ &DSC(\%)$\uparrow$  &JC(\%)$\uparrow$  &HD(pix)$\downarrow$ &ASD(pix)$\downarrow$ \\
		\hline
		U-Net &VGG16 &88.27 &80.04 &10.06 &1.17 &92.16 &86.15 &10.85 &1.19 &34.5\\
		
		DeepLabv3 &Xception &87.71 &79.82 &9.55 &1.24 &91.91 &85.73 &10.21 &1.13 &54.7\\
		
		DeepLabv3  &MixNet &88.46 &80.64 &9.83 &1.07 &92.24 &86.23 &10.82 &1.18 &26.8\\
		
		SPOS          &MixNet &88.32 &80.31 &10.01 &1.14 &92.11 &86.11 &9.92 &1.11 &13.2 \\
		ProxylessNAS  &MixNet &88.55 &80.37 &9.83 &1.13 &92.32 &86.33 &9.83 &1.09 &13.4 \\
		DARTS         &MixNet &88.66 &80.42 &9.55 &1.11 &92.68 &86.42 &9.41 &1.08 &12.9 \\
        Gated-SCNN &WideResNet38 &88.54&80.57&8.68&0.98&92.30&86.21&9.99&1.08&639.1 \\
        Pruning &MixNet &88.13&80.22&10.71&1.31&90.12&85.22&11.21&1.23&13.1 \\ \hline
		
		\textbf{HAS} &MixNet &88.72 &80.85 &9.15 &1.06 &92.72 &86.51 &9.26 &1.04 &\textbf{12.8}\\ 
		
		\textbf{HASA$_{Ac}$} &MixNet &\textbf{89.05} &\textbf{81.26} &\textbf{8.42} &\textbf{0.92} &\textbf{92.75} &\textbf{87.00} &\textbf{8.73} &\textbf{0.91} &\textbf{12.8}\\
		\hline
	\end{tabular}}
	\label{tab:seg}
\end{table*}

The baseline is the DeepLabv3 using MixNet as the backbone. For consistency, all NAS-based experiments shared a similar training scheme. The training data is further separated into a 30\% subset for architecture search, and the remaining 70\% is used to update the network parameters. As shown in Table~\ref{tab:seg}, compared with those state-of-the-art architectures, our proposed method can achieve not only higher performance but also better parameter efficiency. Specifically, for follicle and ovary, the mean DSC of the HASA$_{Ac}$ is improved by 0.59\% and 0.51\% compared with the baseline, respectively. The improvement of the mean JC is 0.62\% and 0.77\%, respectively. This indicates that the segmentation results of the HASA$_{Ac}$ are closer to the ground truth. For HD and ASD which are the smaller the better, the proposed HASA$_{Ac}$ also gets lower errors than the baseline, reducing 1.41 $pixels$, 0.15 $pixels$, and 2.09 $pixels$, 0.27 $pixels$ for follicle and ovary segmentation respectively, which shows that our HASA$_{Ac}$ can achieve better boundary estimation. When compared with the advanced NAS methods, including SPOS, ProxlessNAS, and DARTS, both HAS and HASA$_{Ac}$ present better performance on all metrics with the least parameters. \par

Although applying the pruning approach to the baseline halved the number of the parameters, the performance became worse than that of the baseline. In addition, we compared our method with a gated shape CNN architecture (i.e., Gated-SCNN) \citep{takikawa2019gated}, The results in Table~\ref{tab:seg} show that our proposed method outperforms the Gated-SCNN in all evaluation metrics for segmenting both ovary and follicles. It is also noted that our method has far fewer parameters than the heavy Gated-SCNN. \par

\subsubsection{Examples of image segmentation}
\begin{figure}[ht!]
    \vspace{2mm}
	\centering
	\includegraphics[width=1.0\linewidth]{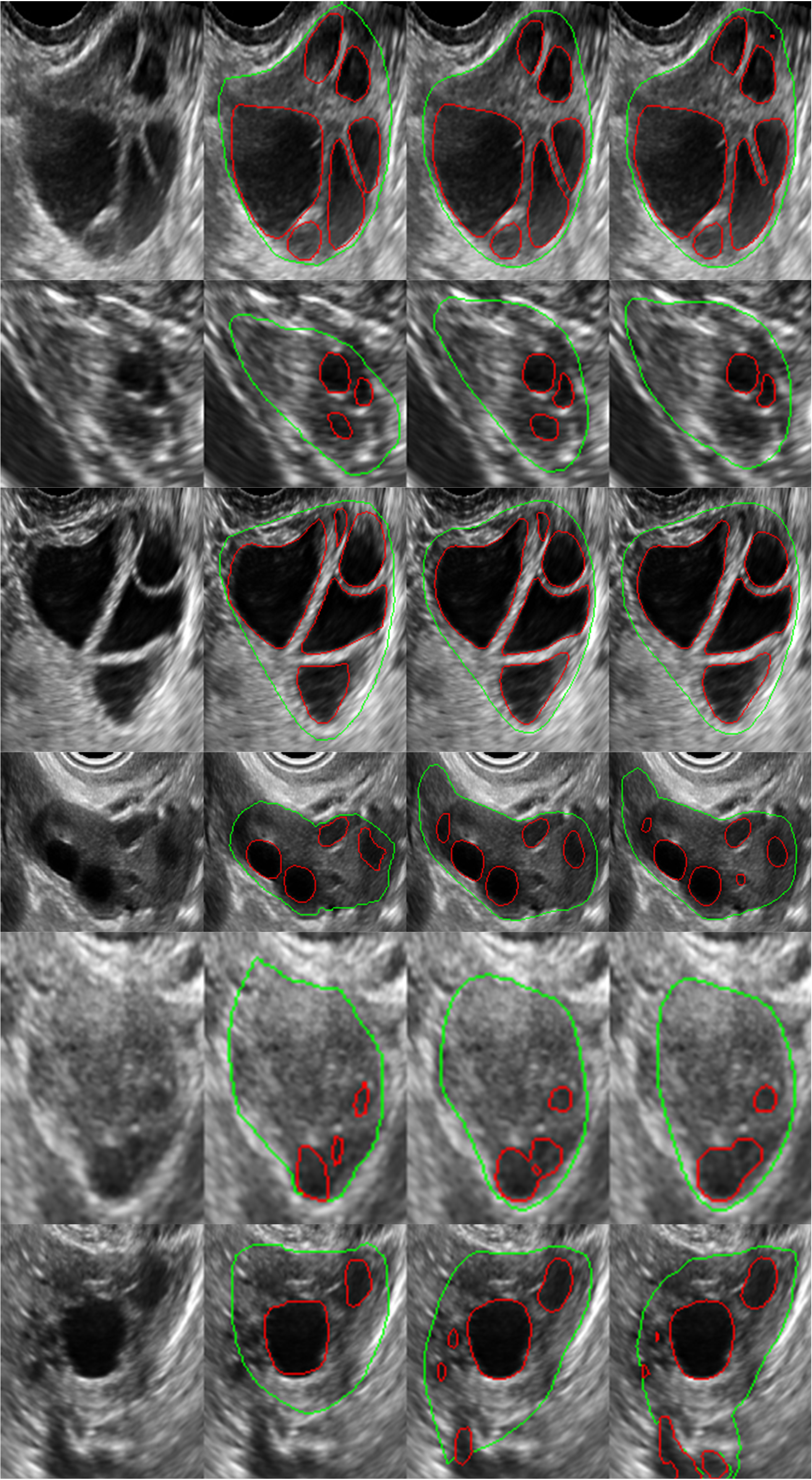}
	\caption{Qualitative comparison of follicle (red color) and ovary (green color) segmentation. From left to right: ovarian US image, segmentation ground truth, result of HASA$_{Ac}$, and result of DeepLabv3. Best viewed in color version.}
	\label{fig:segmentation_result_4}
\end{figure}

Fig.~\ref{fig:segmentation_result_4} qualitatively compares the results of the proposed HASA$_{Ac}$ and the baseline (DeepLabv3 using MixNet as the backbone) on the ovary and follicle segmentation tasks. The images in the first three rows are good examples for both methods. It can be observed that the segmentation results of the HASA$_{Ac}$ are closer to the ground truth than the baseline. Particularly, HASA$_{Ac}$ shows obvious advantages over the baseline in recognizing the tiny follicles (see the second and third rows). The last three rows show three modest segmentation examples. The misrecognition is mainly attributed to the low contrast between the region-of-interest structures and the surrounding tissue. For the image in the last row, both methods tended to misrecognize the hypoechoic areas outside the ovary as follicles due to the very weak signals of boundary. \par

It is worth noting that the proposed HASA costs fewer parameters while improving the segmentation performance. Equipped with the progressive growing search, our HAS can reduce about 52\% parameters when compared with the baseline. This may indicate that our method would be more efficient and economic for the devices with limited computation resources, such as laptop US devices. By further introducing the cell re-aggregation strategy, i.e., ASPP-cell, our HASA can obtain further improvement on all metrics for both the follicle and ovary without adding any extra computation burden. \par

\section{Discussion and conclusions}
Although many hand-designed model architectures have demonstrated great success in various tasks, designing a DNN that works well for specific tasks is still challenging. To avoid the difficulty of manual design, researchers have proposed NAS algorithms that learn high-performance model architectures in an automatic and systematic way.  Despite having this appealing feature, previous NAS studies often suffer from high computational costs due to the massive search space, and the scratch searching scheme which usually needs very large training datasets to achieve good performance. In this study, we aim to develop a lightweight and efficient DNN with high performance. For this purpose, we proposed a hybrid NAS framework with cell re-aggregation strategy and tested its effectiveness on two important tasks: hepatic echinococcosis classification and ovary segmentation in ultrasound images. \par 

Intensive ablation studies and comparisons with other counterparts supported the superiority of our method. The strength of the proposed framework can be mainly attributed to the following three aspects.  First, the hybrid framework combined the pre-trained backbone and NAS cells, rather than search the whole architecture from scratch. This hybrid mode can take advantage of the excellent ability of the pre-trained backbone's initial layers to extract the low-level features, and the search process was only performed for the subsequent NAS cells (Fig.~\ref{fig:framework}). For example, the DarkNet53 architecture contains five stages in which the majority of parameters are in the last two stages (Fig.~\ref{memory}). We replaced the last two stages with NAS cells. Experimental results showed that our method achieved better classification performance with about 40$\%$ fewer parameters compared with DarkNet53 (Table~\ref{tab:class_Table}). \par

Secondly, we introduced the MixConv layer and SE block as the candidate operations, and adopted the progressive growing strategy for the NAS cell search process. Prior NAS studies \citep{liu2018darts, chen2019progressive} mainly used several 2D and 3D vanilla convolution operators. The addition of the two operators enhances the variety of the cell candidates, makes the searched cells more powerful, and thus improves model performance (HAS-w/o-MS vs. HAS in Table~\ref{tab:class_Table}).\par

Thirdly, we exploited expert knowledge in the proposed framework by using expert-designed structures to re-aggregate the searched sequential cells. Conventionally, the searched cells are sequentially stacked. However, such linear structure may have limited feature representation ability. Instead, to re-aggregate the NAS cells, we adopted the dense-cell structure for the classification task and the ASPP-cell structure for the segmentation task (Fig.~\ref{fig:cell_aggregation}). In the dense-cell structure, each cell connects with all subsequent cells, which has multiple advantages such as strong gradient flow and more diversified features. The ASPP-cell structure can incorporate multi-scale context which is vital for image segmentation. Both structures proved to be beneficial to the model performance, as shown in the last two rows in Table~\ref{tab:class_Table} and Table~\ref{tab:seg}. More importantly, the re-aggregation strategy improved  the model performance without adding extra parameters, which is a desirable feature for training DNNs. \par

Although our study provides compelling evidence that the proposed NAS framework is effective in echinococcosis classification and ovary segmentation in ultrasound images, some limitations are worth noting. First, the input of our classification network is the cropped lesion images based on manual annotations rather than the entire US images. To achieve fully automated classification, lesion detection needs to be implemented first. Second, the US images we collected in the study are routine clinical images that were generated by different equipment and different operators. So they may not have similar appearance due to the variations in acquisition, which could have an adverse impact on the subsequent image analysis. Advanced image normalization techniques, such as style transfer and speckle reduction are expected to further improve the performance \citep{roy2020clinical, aysal2007rayleigh}. \par

In this study, we validated the effectiveness of our proposed HASA framework on two large US image datasets for the echinococcosis classification and ovary segmentation tasks. The experimental results showed that our HASA significantly outperforms other competing methods in terms of model size and performance. Our method provides a useful and general tool to automatically generate a lightweight network for image classification and segmentation. Although we only presented here for two specific tasks, this method is readily generalizable to other similar tasks such as the classification of lesions in breast and kidney. We will further extend this work and validate it on higher-dimensional ultrasound images, like 3D ultrasound image analysis. \par

\section*{Declaration of competing interest}
The authors declare that they have no known competing financial interests or personal relationships that could have appeared to
influence the work reported in this paper.\par

\section*{Acknowledgment}
This study was supported by National Natural Science Foundation of China (Nos. 62171290, 61901275, and 62101343), Shenzhen-Hong Kong Joint Research Program (No. SGDX20201103095613036), Shenzhen University startup fund (2019131), Shenzhen Science and Technology Innovations Committee (No. 20200812143441001), and SZU Top Ranking Project (No. 86000000210).\par

\bibliographystyle{model5-names}
\bibliography{HASA_final_revision_clean_version}

\end{document}